\documentclass[conference]{csce}
\usepackage{comment}
\usepackage[american]{babel}
\usepackage[T1]{fontenc}
\usepackage{times}

\usepackage{algorithm}
\usepackage{algorithmic}
\usepackage{mathrsfs}
\usepackage{bm}
\usepackage{color}
\usepackage{slashbox}
\usepackage{setspace}
\usepackage{xspace}
\usepackage{amsmath}
\usepackage{booktabs}

\usepackage{textcomp}
\usepackage{epsfig,graphicx}
\usepackage{xcolor}
\usepackage{amsfonts,amsmath,amssymb}
\usepackage{booktabs}

\title{ \bf 
The application of Evolutionary and Nature Inspired Algorithms in Data Science and Data Analytics}
\author{Farid Ghareh Mohammadi$^1$, Farzan Shenavarmasouleh$^1$,\\ Khaled Rasheed$^1$,Thiab Taha$^1$,   M.  Hadi  Amini$^2$,
 and Hamid R.  Arabnia$^1$\vspace{0.2in}\\  1:  Department of Computer Science,  Franklin College of Arts and Sciences \\ University of Georgia,  Athens,  Georgia,  30602  \\
2:  Knight Foundation School of Computing and Information Sciences\\
Florida International University Miami, Florida 33199\\
Emails :  farid.ghm@uga.edu, fs04199@uga.edu,  khaled@uga.edu, trtaha@uga.edu, amini@cs.fiu.edu,  hra@uga.edu}

\begin{document}

\maketitle                        

\begin{abstract}
In the past 30 years, scientists have searched nature, including animals and insects, and biology in order to discover, understand, and model solutions for solving large-scale science challenges. The study of bionics reveals that how the biological structures, functions found in nature have improved our modern technologies. In this study, we present our discovery of evolutionary and nature-inspired algorithms applications in Data Science and Data Analytics in three main topics of pre-processing, supervised algorithms, and unsupervised algorithms. Among all applications, in this study, we aim to investigate four optimization algorithms that have been performed using the evolutionary and nature-inspired algorithms within data science and analytics. Feature selection optimization in pre-processing section, Hyper-parameter tuning optimization, and knowledge discovery optimization in supervised algorithms, and clustering optimization in the unsupervised algorithms.
 
\end{abstract}

\vspace{1em}

\section{Introduction}
\textit{Motivation}: There are so many nature-inspired and evolutionary algorithms that have been developed based on various natural processes happening in nature. Recently, the usage of these algorithms has proliferated \cite{fan2020review}. Evolutionary and nature-inspired algorithms, such as genetic algorithm (GA), artificial bee colony (ABC), ant colony algorithms (ACO), particle swarm optimization (PSO) have been penetrated in most domains like science according to the high volume of relevant publications \cite{fan2020review, saini2021spam, ch2_farid}. The invention of the technologies and promising results of evolutionary/nature-inspired algorithms to address large-scale optimization problems lead us to drive interest in the evolutionary/nature-inspired algorithms among scientists \cite{fan2020review}.

\begin{figure}[ht]
    \centering
    \includegraphics [width=3in]{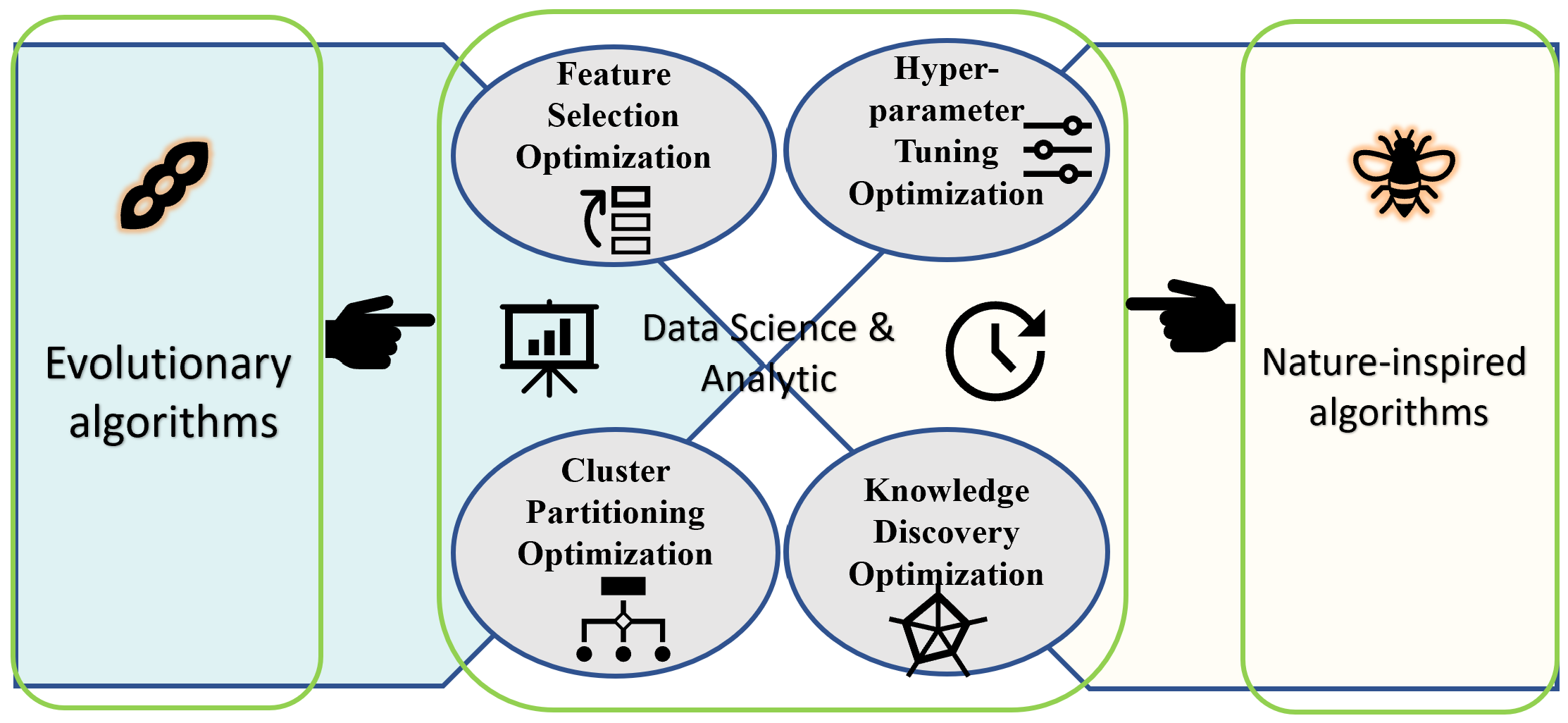}
      \caption{A general schema of Optimization processes in Data science and analytic}

    \label{fig:EV_OPT}
\end{figure}

In this study, we aim to narrow down and be more specific about the applications of the evolutionary and nature-inspired algorithms. Therefore, we plan to provide an overview of applications on four main phases of data science and data analytics: feature selection as a pre-processing algorithm, classification  (hyper-parameter tuning), and knowledge (rule) discovery as supervised algorithms, and clustering as unsupervised algorithms. Figure \ref{fig:EV_OPT} provides a general schema of algorithms and optimizations solutions that have been investigated in this study.


\section{Pre-processing Algorithms}
Pre-processing algorithms entail so many filters and algorithms. Here, we focus on the most important one, feature selection (FS). Additionally, we use FS when we evaluate the classifiers to get a rich model and we use dimension reduction. SO, feature selection may happen in the pre-processing step and classification step.

\subsection{Feature Selection Optimization for Image Classification}
Feature selection is an imperative phase of pre-processing, which affects the performance of classification accuracy \cite{mohammadi2014ifab, kozodoi2019multi}. It generates a subset of features while improving the performance of classification accuracy that is still an NP-hard problem \cite{mohammadi2021evolutionary,ermon2013taming} and scientists have long been trying to solve this problem. Feature selection has been successfully accomplished in several domains   \cite{shi2018efficient, ch1_farid} thus it is not domain-specific algorithm \cite{ch2_farid}.  In this study,  we discuss some certain important feature selection algorithms such as and  IFAB  \cite{mohammadi2014ifab}. 
Kozodoi \emph{et. al.}   \cite{kozodoi2019multi}  used profit measures to propose a solution for wrapper-based feature selection optimization namely, Non-dominated Sorting-based Genetic Algorithm (NSGA-II), which is a non-dominated sorting-based genetic algorithm using a fitness function called Expected Maximum Profit (EMP) measure to calculate the model profitability to select the best features.

IFAB, stands for Image steganalysis usingFeature selection based on ABC, and is evaluated on different sets of digital images like BOSS, the breaking out steganography system \cite{bas2011break} version1.01 greyscale image databases in which texts are embedded with a rate of 0.4 per pixel. This database entails 10,000 cover and stego images each.

\begin{figure}[H]
    \centering
    \includegraphics[height=1.79in]{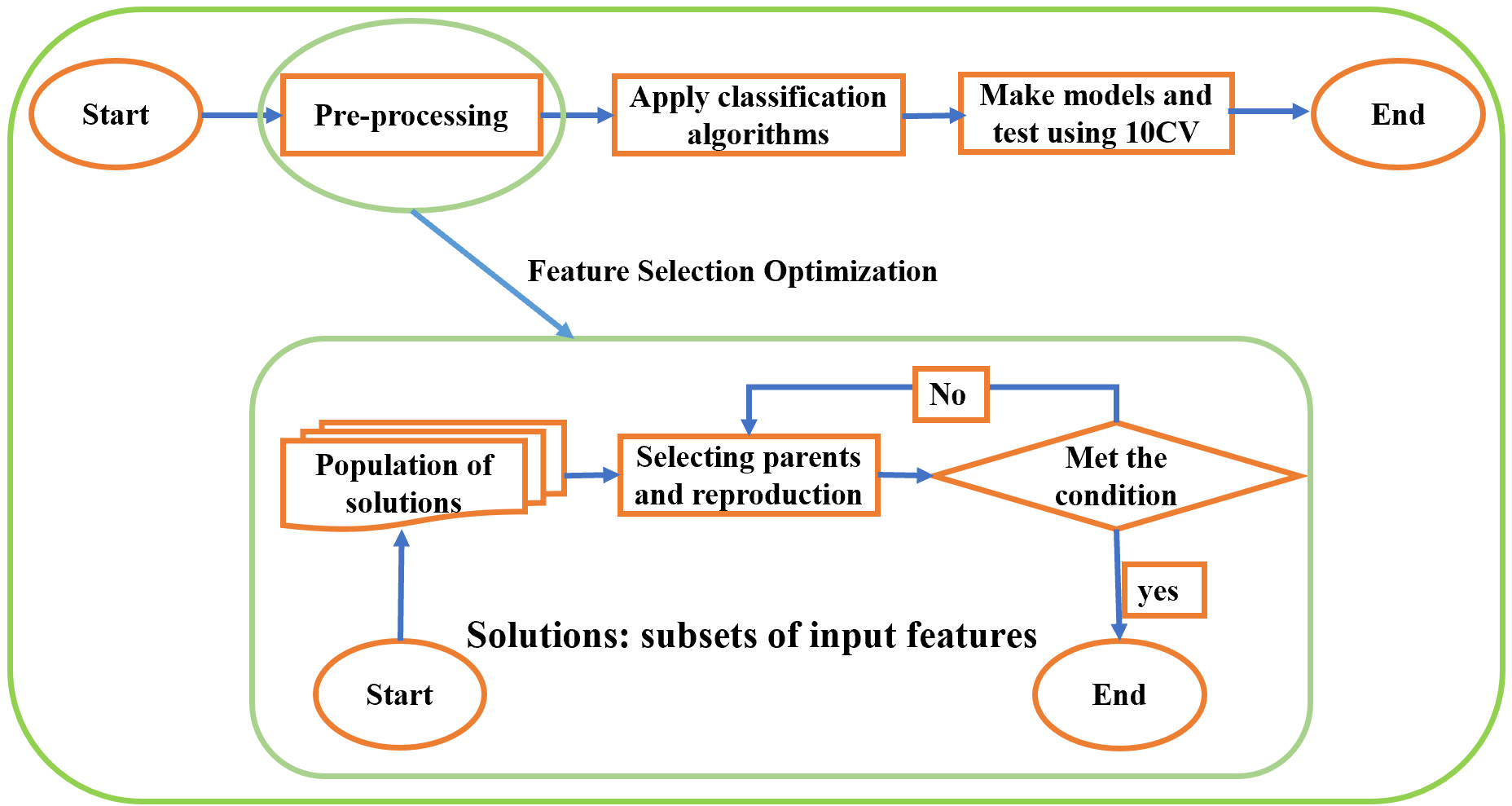}
      \caption{Evolutionary/nature-inspired algorithms process for  feature Selection. (extended and improved from our work in \cite{ch2_farid} }

    \label{fig:EV_FE-general}
\end{figure}

In figure\ref{fig:EV_FE-general}, we present a very abstract view of FS process using evolutionary / nature-inspired algorithms borrowed from \cite{ch2_farid}. This figure shows that Evolutionary algorithms (EAs) are used in pre-processing step in which they enable scientists to choose the best features. To do so,  we use one of the classifications algorithms to learn from a training dataset and generate a rich model using the selected features. Then,  we leverage the generated model to predict the unseen test dataset and evaluate the classifier's accuracy. Note that solutions in the population of food sources are the subsets of features with a designated size. Every time this solution gets updated with the best solution. In other words, the solution with the minimum goodness is replaced with a new solution that gets goodness better than the parent solution.
If we do not consider the best features, classifiers may fail to learn from a large amount of data due to the CoD machine learning problem. So, speaking of feature selection optimization, we recommend an optimal way that classifiers to learn from the training dataset without struggling with over-fitting or under-fitting problems by preventing from learning from irrelevant features.

Generating images in different areas have proliferated, thus image classification became a popular task among scientists. To do multi-class classification,  scientists have developed a large number of packages such as deep learning   \cite{khan2019novel}, convolutional neural network (CNN) \cite{zhang2019sar}.
\setlength{\parindent}{0.5cm}

These developed algorithms have been successfully performed and yielded a very high accuracy in comparison with the traditional machine learning algorithms. However,  they still are struggling to handle big data, particularly CoD problem and their main disadvantage is that deep learning algorithms take a long time to get the training phase done, so they have a high time complexity. Thus, researchers have put their effort to take advantage of evolutionary/ nature-inspired algorithms to address these problems properly. Image classification has been one of the emerging challenges in large-scale science and computer vision problems \cite{mohammadi2017region}.

\subsubsection{Application of Artificial Bee Colony in Feature Selection}
ABC has many applications in the world of data science such as IFAB. IFAB, is an image steganalysis approach using feature selection based on artificial bee colony is presented in \cite{mohammadi2014ifab}.  IFAB is one of the optimum wrapper-based feature selection processes. This method works best for evaluating the generated features from images in order to distinguish cover images from stego images. A stego image is a cover image with an embedded message. The goal of IFAB is to improve the classifier's accuracy by reducing the time complexity of the training model and reducing the feature dimension.

IFAB \cite{mohammadi2014ifab}, as one of the nature-inspired algorithms, entails three types of bees to boost the feature selection process. First,  employed bees are used to calculate a fitness using \ref{eq.fitness} by performing the goodness of each food source using support vector machine(SVM). A solution presents the number of features that have already been selected to decrease the number of features in the given dataset.

\begin{align}
   Fitness (f_i)=
\begin{cases}
    \frac {1}{1+P_i} & \text{if } P_i > 0\\
         1+abs(P_i) & \text{if } P_i <  0 \label{eq.fitness}
\end{cases}
\end{align}

The second step is to select the best food source based on equation \ref{eq.SelectFS} the goodness of the food source to exploit it by onlooker bees. Once offspring (new solution) is evaluated,  we check the limit parameter if the scout bee is needed or not.

\begin{equation}
{V_i}= {f_i}+{v * (f_i-f_j)}. ,  v =[-1, 1]
 \label{eq.SelectFS}
\end{equation}

Where j is a random number generated from 1 and N. i, j are features index in the given dataset. Furthermore, N is the upper bound of the number of features. If the condition for limit is met then,  in the third step,  Scout bee is selected ( it was an employed bee who abandoned a food source to explore using equation \ref{eq.Scout} the space of the food source,  and update the solution by a pre-adjusted rate. $X_max$ and $X_min$ denote the upper bound and lower bound of population,  respectively.
\begin{equation}
{X_i}= {X_{max}}+{v'} * { ({X_{max}}-{X_{min}})} ,  v' =[0, 1] 
\label{eq.Scout}
\end{equation}

Within each iteration, the best food source is recorded and ABC replaces the minimum food source with the new food source if its goodness outperforms the selected food source. ABC terminates if it meets the termination condition or maximum iteration is already completed. In the end,  ABC passes the best food source with the highest fitness value. This food source entails the number of imperative features which yield the best accuracy.


IFAB-KNN is an extended ABC for image steganalysis to enhance IFAB accuracy \cite{mohammadi2014IFABKNN}. IFAB-KNN denotes another wrapper-based feature selection algorithm in which ABC takes advantage of a lazy algorithm,  k-Nearest Neighbor(KNN),  that enables ABC to evaluate the goodness of each food source. Having the same number of selected features with IFAB,  IFAB-KNN outperforms IFAB with the tuned hyper-parameters.

Ghareh mohammadi and Saniee Abadeh \cite{mohammadi2017RISAB} presented another hybrid approach for feature and region (sub-image) selection, using a combination of data and images. They proposed RISAB, region-based image steganalysis using artificial bee colony using IFAB results. The goal of RISAB is to find the best location (sub-image or region) of the given image that does not follow the harmony of the whole images. Seeking special pixel or sets of pixels,  RISAB can differentiate stego images from cover ones. In RISAB,  the researchers first applied ABC to explore the sets of sub-image that involve the embedded information,  which would be images, messages. After that,  there is one given input image, together with the associate sub-image. Then,  researchers evaluated IFAB on both images. RISAB improves the accuracy of prediction of feature extractors even IFAB.

\subsubsection{ Application of Particle Swarm Optimization (PSO) in Feature Selection}
Chhikara \emph{et. al.}   \cite{chhikara2016hybridPSO} proposed a new HYBRID approach leveraging PSO as a solution for a CoD problem in image classification. This hybrid approach combined both filter and wrapper-based feature selection approaches to solving the high dimensionality or the COD problem in image steganalysis. The HYBRID boosted the accuracy of image classification. Furthermore, Chikara and Kumari in   \cite{kumari2017GLBPSO} introduced a new wrapper-based feature selection for image classification,  named Global Local PSO (GLBPSO) which takes advantage of backpropagation neural networks to calculate the goodness of the selected feature subsets. 

Adeli and Broumandnia in   \cite{adeli2018imagePSO}  investigated a new filter-based feature selection algorithm for steganalysis titled,  an Adaptive inertia weight-based PSO (APSO). Additionally, Rostami and Khiavi  \cite{rostami2016PSO} used a new fitness function using Area Under Curve (AUC) to evaluate selected features. The accuracy result of APSO shows that it has a better result in comparison with IFAB.

\subsubsection{Application of Grey Wolf Optimizer (GWO) in Feature Selection} 
Pathak \emph{et. al.}   \cite{pathak2019GWO} proposed LFGWO, which is levy flight-based greu worl optimization, as a new feature selection algorithm that has been utilized to classify stego images from cover images. GWO has been highly applied to solve large-scale optimization problems,  such as  PSOGWO is presented in \cite{al2019binary} an optimal feature selection algorithm using a combination of two algorithms GWO and PSO.

\subsection{Feature Selection Optimization for Network Traffic Classification}
Internet traffics growth has proliferated because of expanding new technologies such as the internet of things (IoT) \cite{ghareh2021data, shenavarmasouleh2021embodied}. Having multi-class datasets with imbalanced class labels become an emerging challenge for scientists. ML algorithms struggle to analyze the datasets and do not perform a high recall for the minority classes. Researchers in their work has investigated hybrid solutions to solve these problems. Dong \emph{et al}  \cite{dong2017efficient} presented a new solution using PSO to solve the challenges. Researchers in this paper used a hybrid feature selection algorithm with a combination of two algorithms: RelieF and PSO algorithms namely, RFPSO. RFPSO works with two main steps. The first step denotes the initialization of RFPSO,  and the second step is fitness function selection using a PSO.

Shi \emph{et. al.}   \cite{shi2017efficient,  shi2018efficient} introduced a solution to classify traffic data by leveraging a combination of feature extraction and selection algorithms in which the solution used a PCA. Furthermore, Hamamoto \emph{et. al.}   \cite{hamamoto2018network} leveraged a Genetic algorithm (GA) to present a network anomaly detection solution. The GA is used to make a digital signature of the network segment using flow analysis. The required information is extracted from the network flows which have been utilized to predict networks traffic action. They combined Ga with a fuzzy logic algorithm to enhance the output result. The fuzzy logic algorithm enables the Ml algorithm to make a decision if a new instance is an anomaly or not.

\subsection{Feature Selection Optimization Benchmarks}
In this section, we discuss the further discovery of feature selection optimization in the world of data science and analytics. Ant colony optimization attribute reduction (ACOAR) \cite{ke2008efficient} as a filter-based feature selection algorithm reducing the number of features by applying ACO to enhance the performance of the algorithm in rough set theory. Additionally, UFSACO, an unsupervised feature selection method based on ant colony optimization,  is proposed in \cite{tabakhi2014unsupervised}  which is a novel feature selection algorithm for unsupervised data using similarity of features. In other words, the goodness of features relation is computed with respect to the similarity among features. However,  it does not use any classifiers to perform feature selection.

\begin{figure*}[H]
    \centering
    \includegraphics[height=1.79in]{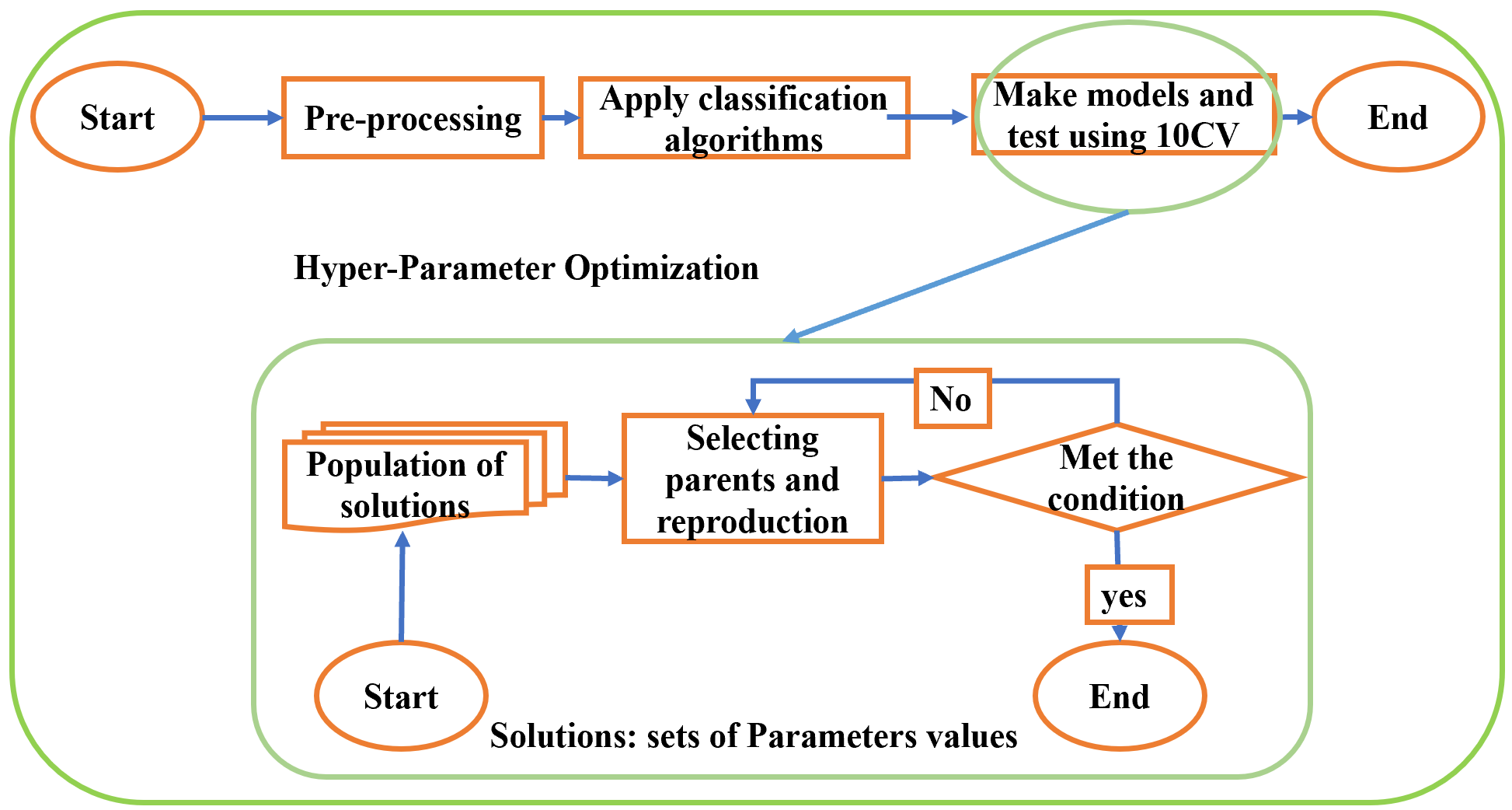}
      \caption{Evolutionary/nature-inspired algorithms process for Hyper parameters Optimization
      }

    \label{fig:EV-HPT-general}
\end{figure*}

\section{Supervised Algorithms}

Classification is the main task of supervised algorithms in which classifiers predict class labels to previously unseen instances in the test dataset. To do the classification task, we take an action by training a classifier, then this classifier is evaluated on unseen instances in the test dataset to assign class labels \cite{zorarpaci2021privacy}.
 
Classification has diverse application domains from image classification to identify certain objects\cite{mohammadi2014ifab,mohammadi2017region}, from text classification to identify its author, and sentiment analysis to identify the type of the text \cite{osmani2020sentiment}. Thus, we need to train classifiers for each dataset and domain every time from the beginning. It means that we cannot apply a classifier that is already trained for a certain category of domains to another dataset and domain. Therefore, for each dataset, we need to have a separate classifier that is trained. In order to have the classifier to get trained well to yield promising results, we need to let the classifier make a rich model learning from the training dataset. 

A classifier can get a rich model by getting customized and tuned by adjusting the hyper-parameters. Furthermore, in order to get the best model for rule-based classifiers, we require to have an optimized solution to extract rules and knowledge so that enables classifiers to yield the best results. Here, we discover the applications of evolutionary and nature-inspired algorithms for hyper-parameter optimization and knowledge discovery optimization.

\subsection{Hyper-parameter Optimization}
All supervised algorithms, particularly classification algorithms model entails parameters that are needed to get tuned for making a rich model. The parameters that are required to get initialized so that let the algorithm run is called Hyper-parameters (HPT) \cite{zahedi2021search}. Note that HPT values will not change during the training process. In contrast, some parameters get initialized and updated during the training process \cite{kuhn2013applied}.
 
Tuning the HPs in a way that classifiers can yield the best performance in terms of accuracy is defined ad automated ML (AutoML) \cite{zahedi2021search, feurer2019hyperparameter}. There are two popular basic automated HPO methods: grid search  (GS) and random search (RS). However, these two algorithms take longer than we expect to tune the machine learning algorithms parameters \cite{zahedi2021search}. For some classifiers, it takes around one year to get tuned \cite{zahedi2021search}. Thus, we need evolutionary/ nature-inspired algorithms to optimise this process \cite{ch2_farid}. Figure \ref{fig:EV-HPT-general} presents an abstract procedure of hyper-parameters tuning that performs within a classifier leveraging one of the evolutionary/nature-inspired algorithms. The algorithms take care of tuning the hyper-parameters in a way that the classifier yields the best accuracy ever in which evaluated on a dataset. Each solution in the population of food sources is a set of values for the hyper-parameters per classifier. At the end of the process of hyper-parameter tuning, the best set of values associated with the hyper-parameters are recorded then the classifier sets the configuration accordingly and yields the best result.

\begin{figure}[H]
    \centering
    \includegraphics[height=1.79in]{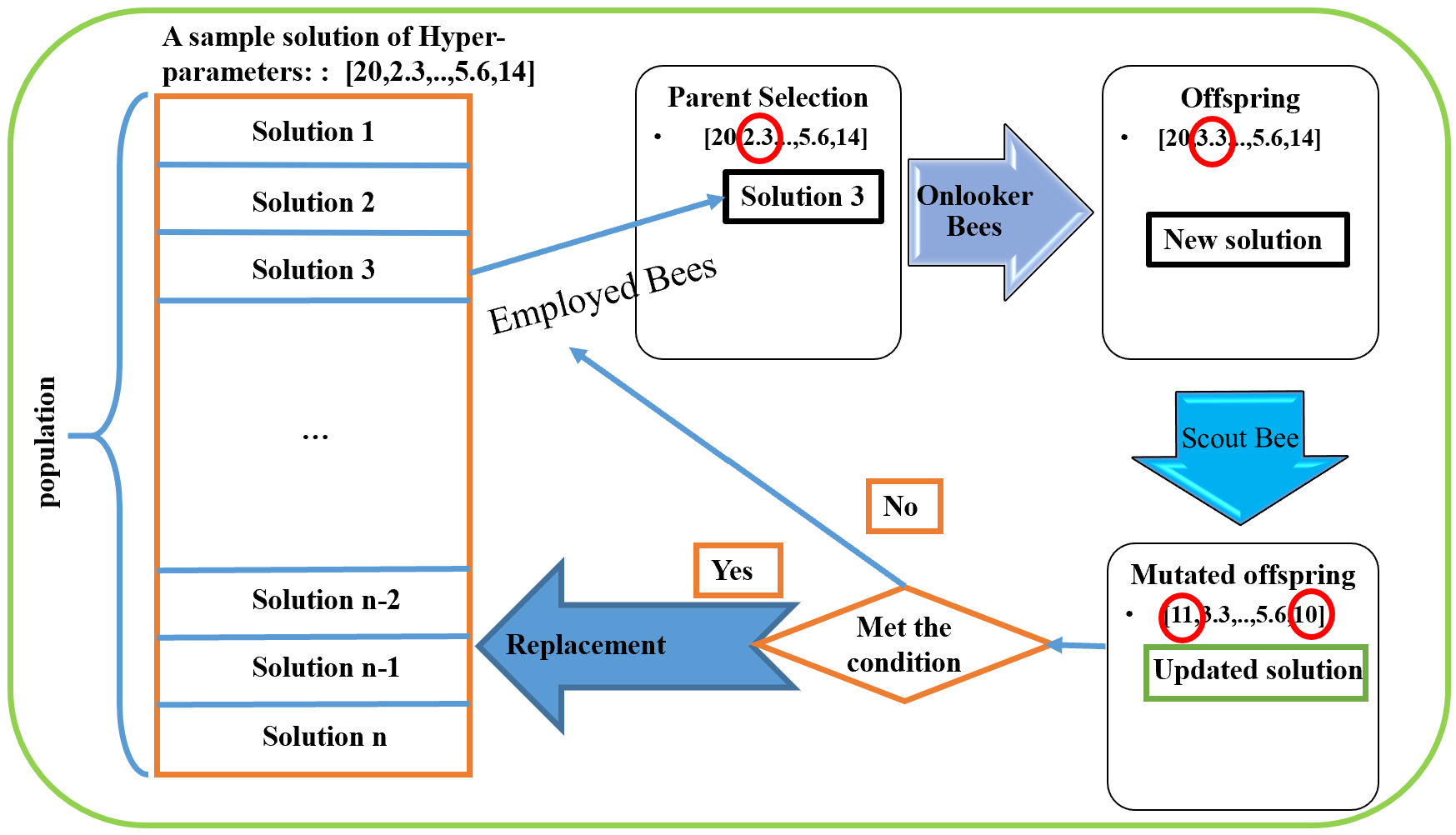}
   
      \caption{A general Schema of Hyper-Parameters tuning by ABC
      }

    \label{fig:HPT-ABC}
\end{figure}
In the literature, we discover that classifiers such as SVM, MLP leverage evolutionary/nature-inspired algorithms like ABC algorithm to optimize their parameters \cite{zahedi2021hyp} , ABC has also been used to design automatically and evolve hyper-parameters of Convolutional Neural Networks (CNNs) \cite{zhu2019evolutionary}. Figure \ref{fig:HPT-ABC} shows a detailed process of ABC in which the algorithm chooses a solution consisting of a set of values for hyper-parameters and updates the solution using onlooker bees and uses a scout bee to randomly choose another solution that has not already been discovered yet.

\begin{figure}[H]
    \centering
    \includegraphics[height=1.79in]{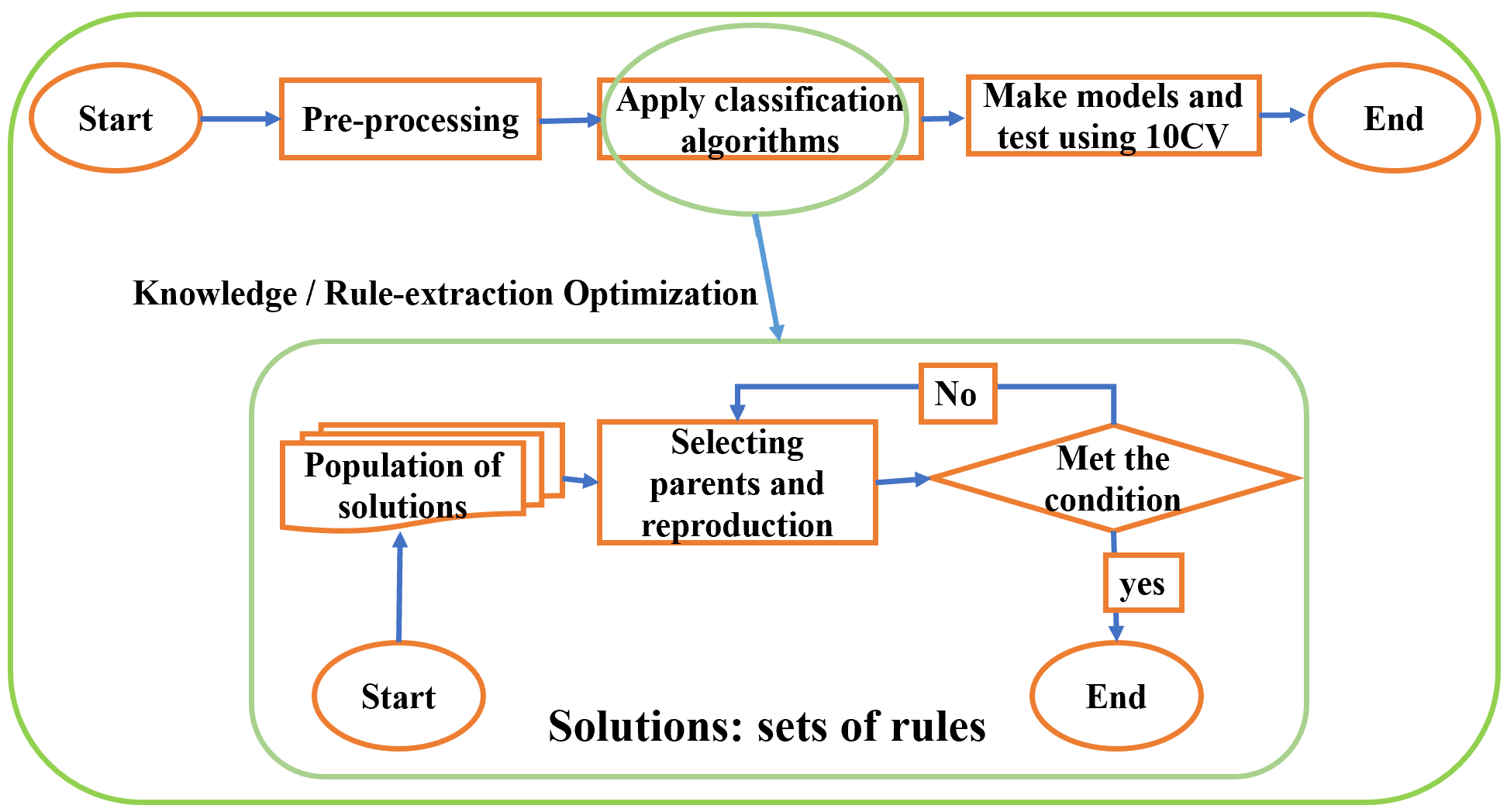}
      \caption{Evolutionary/nature-inspired algorithms process for Knowledge and Rule extraction Optimization
      }

    \label{fig:EV-KRE-general}
\end{figure}
\begin{figure}[ht]
    \centering
    \includegraphics[width=3in] {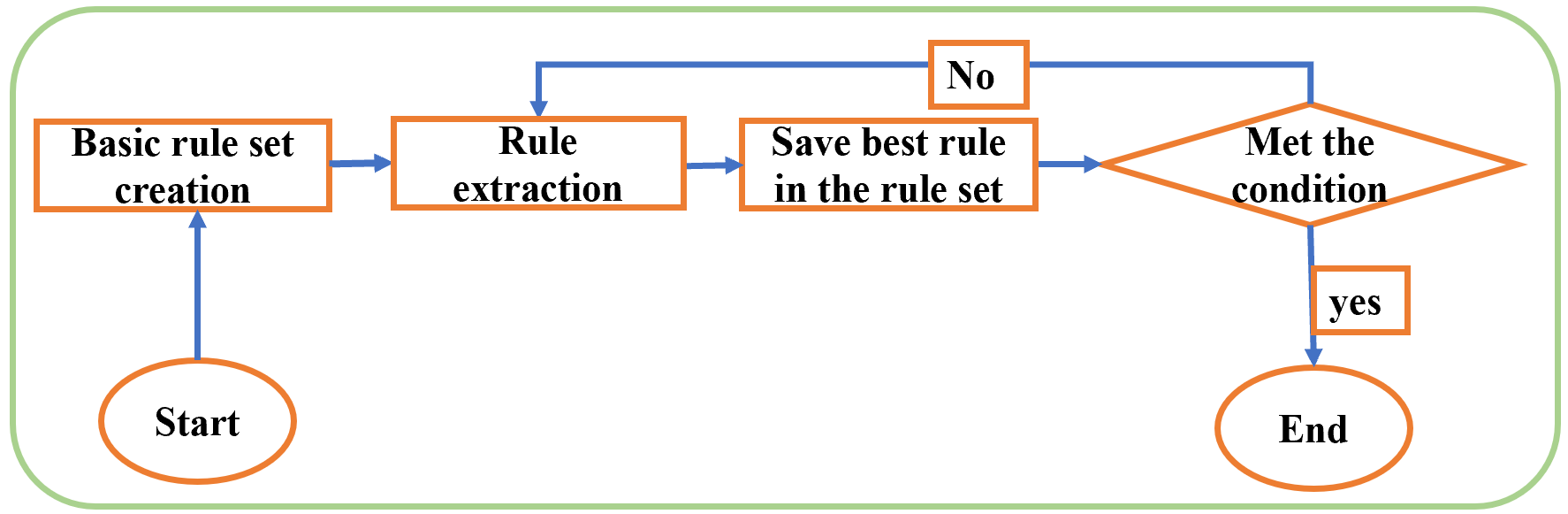}
   
      \caption{Knowledge and Rule extraction Optimization
      }
    \label{fig:KRE-optimization}
\end{figure}
\subsection{Knowledge and Rule Discovery Optimization}
Knowledge Discovery optimization has been growing recently using different approaches such as evolutionary algorithms \cite{llorente2021knowledge} and nature-inspired algorithms \cite{zorarpaci2021privacy}.  Researchers in \cite{llorente2021knowledge} presented a new solution for knowledge discovery namely, hybrid model of Genetic algorithm and Genetic programming(GP), GA-GP, using a combination of genetic algorithm and genetic programming that extracts rules and knowledge in the given dataset using compensatory fuzzy logic (CFL). With the growth of generated datasets, the automated discovery of knowledge in the given datasets has become imperative to do proper analysis and make an optimal decision \cite{llorente2021knowledge}. 

We basically focus on ABC to state the importance of this algorithm in an optimization process.  AS artificial bee colony have been used to discover classification rules lately and the success of ABC algorithm for the discovery of classification rules has been investigated \cite{zorarpaci2021privacy}. Thus, In this section, we review recent applications of evolutionary and nature-inspired algorithms in knowledge and rule discovery. Figure \ref{fig:EV-KRE-general} depicts the general overview of a Knowledge and rule discovery process. In this figure, we show how the knowledge/rule discovery optimization step occurs within the typical data science / analytic procedure to boost classification accuracy. The population of the food source entails the rules which are considered as solutions to be selected to get updated and enable the classifiers to yield a better result.

Figure \ref{fig:KRE-optimization} states a detailed process of knowledge discovery optimization. First, we generate rule sets as many as possible. Second, we extract the possible rules that cover the majority of instances in the given dataset. Third, we search to find the best rule set that yields the best result of classification accuracy, and then we record the ruleset. Fourth, we check if the condition is met or not. The condition would be the length of the ruleset and/or the accuracy of the classification and/or the max iteration is completed. If so, we choose the ruleset to configure the classifier to get evaluated on the test dataset. If not, we need to repeat steps from steps 2 to 4.

\subsubsection{Applications of Artificial Bee Colony }
In addition to classification and clustering, we have a knowledge/rule-extractor optimization approach so that enables scientists to have better rule-based algorithms.  
One of the earliest studies is ABCminer presented in \cite{celik2011artificial}. It takes advantage of ABC algorithm to extract classification rules from the training dataset. ABCminer is evaluated on the discovery of classification rules over the datasets such as zoo, Breast Wisconsin, and Breast Tissue. 

In 2014, Talebi and abadi have constructed a new rule-extractor named BeeMiner \cite{talebi2014beeminer} which is a modified ABC algorithm for the aim of rule discovery. BeeMiner applies an information-theoretic heuristic function (IHF) rather than using the basic ABC algorithm for seeking areas across the search space. Additionally, Cooprative artificial bee colony miner (CoABCMiner) is a cooperative algorithm \cite{celik2016coabcminer} has been introduced for a rule-based classification algorithm in which it extracts all classification rules at once. In CoABCMiner, ABC algorithm performs to discover and extract all classification rules and learn from the rule list. It has been stated that CoABCMiner is evaluated for the discovery of classification. 

In \cite{zorarpaci2021privacy} researchers proposed a new solution of rule-discovery optimization algorithm for the supervised algorithms using ABC and perturbation technique of differential privacy to execute privacy-preserving classification. This solution works based on a two-tier solution structure and uses binary optimization operators to identify if a feature is in or not in the conditions part of an IF-THEN rule in the first-tier solution structure. In the second tier, the researchers aimed to use a combination of mutual support and logical sufficiency, which are the two quality measures. According to the results reported in \cite{zorarpaci2021privacy}, the solution ABC and rule-based classifier outperforms typical machine learning rule-based algorithms such as PART, support vector machine (SVM), Holte’s One Rule, and C4.5 over public and/or private supervised datasets.

\section{Unsupervised Algorithms}
K-means clustering algorithm, as one of the unsupervised algorithms, calculates the correct combination of centroid positions for every category in each step. When data points look simple and no certain formation is found, the generated centroids are locally optimum in their intra-categories and global optimum in their inter categories where the properties state maximum / minimum similarities among categories in clustering.

K-means performs the clustering properly and provides the following advantages. It is relatively simple to implement. Second, it guarantees a convergent process.  Third, it readily adapts to new instances. However, generalizing to clusters of different structures, shapes, and sizes, such as elliptical clusters and examining outliers looks challenging issues. One major drawback of K-means clustering is the lack of global optimum in which the starting
partition is selected using random centroid values without guaranteeing the best possible clustering result\cite{shaik2020survey}.

Evolutionary and nature-inspired optimization algorithms can address drawbacks of the K-means algorithm effectively that are computed randomly from beginning to its convergence \cite{mohammadrezapour2020fuzzy}. Researchers in \cite{mohammadrezapour2020fuzzy} presented two different algorithms one c-means with fuzzy logic and one K-means with genetic algorithm.  The algorithms seek the agents based on centroids principal that enable the new clusters to be generated to yield the optimum setting of centroids using the following formula borrowed from \cite{shaik2020survey}
:
 
\vspace{0.1in}
$W_{i, j}=\{0, x_i \notin cluster j | 1, x_i \in cluster j \}$     \quad        \quad (1)

\vspace{0.2in}
Where $W_{i, j}$ is the representatives of data point $X_{ij}$, and i,j are clusters.
\vspace{0.1in}

$centeoid_{i, j}= \frac{\sum_{i=0} ^{s} w_{i, j} x_{i, v}}{\sum_{i=0} ^{s} w_{i, j}}$,   \quad  \quad \quad  \quad  \quad  \quad\quad  (2)

\vspace{0.09in}
 \textit{ $j=1,... K, v=1,..., K * D$  } 
\vspace{0.2in}

According to the equation (2) \cite{shaik2020survey}  S is the number of search agents in the whole population, K denotes the maximum number of clusters and j is the current cluster within the dataset with a size of D.

\subsection{Bee Optimization Algorithms}
There are different versions of Bee optimization that we found two of them more popular and used in the context of this study. Bee inspired optimization algorithm and Artificial Bee Colony (ABC). Saini \emph{et al} in \cite{saini2021spam} presented a customised ABC to perform Kmeans. Algorithm \ref{ABC-Kmeans} presents a pseudo-code of ABC kmeans for cluster head optimization. This algorithm was able to perform clustering using ABC consisting of three main sections each assigned to three types of bees: Employed bees, Onlooker bees, and one scout bee. Finally, the algorithm stores the best agent and returns optimal cluster heads.

 \begin{algorithm}[H]
  \caption{Implementation of ABC  Kmeans for spam clustering (cluster head optimization) \cite{saini2021spam}}
  \begin{algorithmic}[1]
  \renewcommand{\algorithmicrequire}{\textbf{Input:}} 
  \renewcommand{\algorithmicensure}{\textbf{Output:}}
   \REQUIRE data, The number of cluster
  \ENSURE  Best agent / Optimal cluster heads\\
  \STATE {Initialize the search agents with random values and then calculate their fitness}
  \WHILE{Stopping criteria is not met}
  \STATE {Employed bees select the search agents(each per employed bee) update the agent and recalculate the fitness}
  \STATE Onlooker bees select a search agent having best chance to update the solution
  \STATE Scout bee reinitialize the search agent sources randomly and calculate the fitness
  \STATE Memorize and update the best agent
  \ENDWHILE
  \RETURN Best agent / Optimal cluster heads
  \end{algorithmic}
  \label{ABC-Kmeans}
  \end{algorithm}

\section{Conclusion}
In this study, we provide a review of the applications of evolutionary and nature-inspired algorithms in data science and data analytics. We investigate the applications in three main sections of pre-processing and supervised algorithms and unsupervised algorithms. We discuss four important optimization solutions for the given sections. We present one pre-processing algorithm, feature selection optimization, two supervised optimizations algorithms. The first one is Hyper-Parameter tuning optimization and the second one is knowledge or rule discovery algorithms. Additionally, we depict how nature-inspired algorithms have impacted the optimization process of clustering algorithms.

\bibliography{bib.bib}
\bibliographystyle{IEEEtran}

\end{document}